\documentclass{article}
\usepackage{spconf,amsmath,graphicx,hyperref}
\usepackage{array,tabularx,booktabs} 
\newcolumntype{Y}{>{\centering\arraybackslash}X} 
\usepackage{enumitem}

\usepackage{hhline} 

\title{Analysis of Long Range Dependency Understanding in State Space Models}

\name{Srividya Ravikumar$^{*}$ \qquad Abhinav Anand$^{*}$ \qquad Shweta Verma$^{*}$ \qquad Mira Mezini$^{* \dagger \ddagger }$}
  
  \address{$^{*}$ TU Darmstadt \\
      $^{\dagger}$ Hessian Center for Artificial Intelligence, Darmstadt, Germany \\
     $^{\ddagger}$ National Research Center for Applied Cybersecurity ATHENE}

\begin{document}
%
\maketitle
\begin{abstract}
Although state-space models (SSMs) have demonstrated strong performance on long-sequence benchmarks, most research has emphasized predictive accuracy rather than interpretability. In this work, we present the first systematic kernel interpretability study of the diagonalized state-space model (S4D) trained on a real-world task (vulnerability detection in source code). Through time and frequency domain analysis of the S4D kernel, we show that the long-range modeling capability of S4D varies significantly under different model architectures, affecting model performance. For instance, we show that the depending on the architecture, S4D kernel can behave as low-pass, band-pass or high-pass filter. The insights from our analysis can guide future work in designing better S4D-based models.  
\end{abstract}
%
\section{Introduction}
\label{sec:intro}
Structured State Space Models (SSMs) \cite{lssl} have recently been introduced as efficient architectures for sequence modeling, showing competitive results in language, time series, and program analysis tasks \cite{verma2025codessm}. By parameterizing input–output mappings through long convolutional kernels, they provide a scalable way to capture long-range dependencies. The original S4 model demonstrated this potential on long-sequence benchmarks, positioning SSMs as strong alternatives to RNNs and transformers. The diagonal variant of SSM, S4D \cite{gu2022parameterizationinitializationdiagonalstate}, further improved computational efficiency and stability by simplifying kernel generation, making it more practical for large-scale applications. Recent work on understanding source code has explored both transformer-based approaches and state-space variants, with CodeSSM \cite{verma2025codessm} indicating that SSMs can effectively model the syntactic and semantic structure of code.

Despite these advances, the mechanisms underlying the behavior of SSM remain insufficiently understood. Previous research has examined stability and generalization properties of SSMs (see Section~\ref{priorwork}), but the practical implications for kernel behavior are less clear. In particular, it is not well established whether kernels emphasize local transitions, long-range dependencies, or specific spectral bands, nor how these properties vary across architectures. This lack of interpretability limits our ability to connect theoretical predictions with empirical outcomes.

In this work, we study the interpretability of SSM kernels using the diagonalized variant S4D, which provides computational efficiency and stable kernel generation while preserving the modeling capacity of SSMs \cite{gu2022parameterizationinitializationdiagonalstate}. We analyze CNN–S4D hybrid architectures in both the time and frequency domains to examine how architectural choices shape kernel responses and influence the modeling of dependencies. For our study, we used single-layer models to maintain architectural simplicity and interpretability, enabling detailed analysis of how individual kernels respond to specific code patterns. We trained the models on a real-world task (vulnerability detection in source code) instead of synthetic datasets used in previous works on SSM analysis. Single layer 1D CNNs have already been shown to perform well on this task. So, we use a single-layer CNN baseline to evaluate the performance of different S4D-based models.

Our observations depict that the standalone S4D model performs worse than CNN. Consistent with this observation, our analysis of learned S4D kernel reveals that standalone S4D model only learns short-range dependencies, despite its theoretical long-range modeling capacities \cite{lssl, gu2022efficientlymodelinglongsequences}. On the other hand, the S4D model preceded by a state memory replay \cite{qi2024smrstatememoryreplay} block performs the best and also shows long-range understanding in our analysis.

\section{METHODOLOGY}
In this section, we explain the different model architectures we have trained and the proposed methodology to analyze the behavior of the SSM kernel. 

\subsection{Model architectures}

\begin{figure}[t]
    \centering
    \includegraphics[width=0.96\linewidth,trim={1cm 7cm 0cm 7cm},clip]{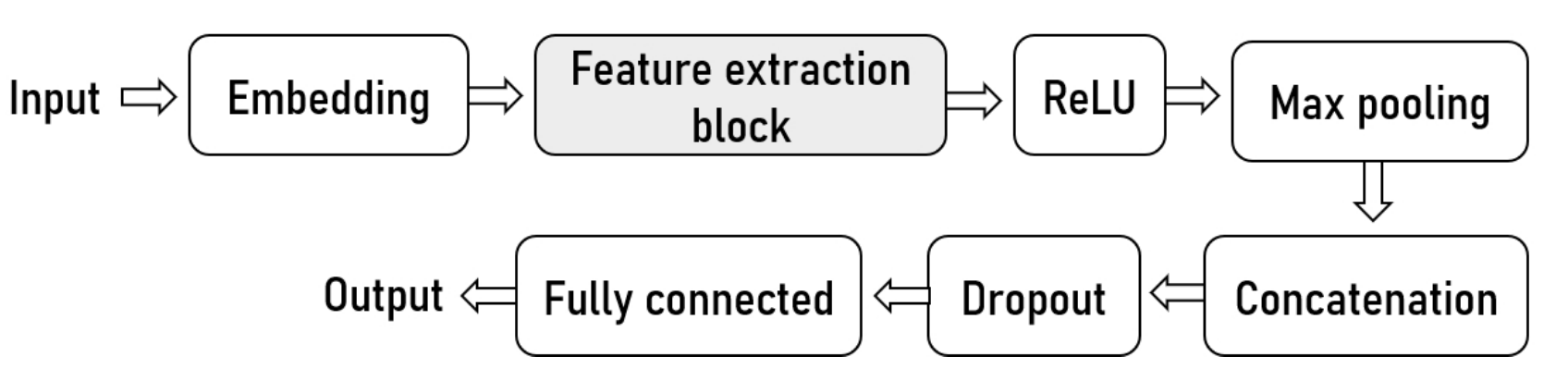}
    \caption{Common model architecture used across all experiments. The Feature extraction block represents different feature transformation strategies across six model variants.}
    \label{fig:common arch}
\end{figure}

We implemented six model architectures to evaluate performance and study kernel behavior in vulnerability detection. All models share a common architecture (Figure \ref{fig:common arch}), where tokenized source code is mapped to dense embeddings and processed through a feature extraction block. The outputs are passed through ReLU activation, max pooling, concatenation, dropout, and a fully connected layer for binary classification. The feature extraction block is the only component that varies between models.
    
    \textbf{Convolution 1D baseline model:} A single one-dimensional convolutional layer that applies multiple kernel sizes in parallel to extract local features similar to CNN architectures used in sentence classification tasks \cite{kim2014convolutionalneuralnetworkssentence}.
    
    \textbf{Depthwise separable convolution model:} The model factors the standard convolution into two successive stages: a convolution in depth, where a single filter is applied per channel to extract the localized structure, and a pointwise (1×1) convolution, which mixes information between channels significantly reducing the number of parameters and computational cost while maintaining predictive performance \cite{chollet2017xceptiondeeplearningdepthwise}.
    
    \textbf{S4D (standalone) model:} In this architecture the feature extraction block contains a single Structured State Space (S4D) layer. The S4D efficiently computes parameterized kernels in the frequency domain using FFTs.
    
    \textbf{Depthwise separable with S4D model:} This hybrid design first applies a depth-wise convolution for local feature extraction, followed by an S4D layer with global context modeling, and finally a point-wise convolution to integrate information across channels. This model was designed to illustrate how SSMs and CNNs can complement each other in capturing both structural and semantic cues in code.
    
    \textbf{Conv1D + S4D model:} This hybrid model extends the Conv1D baseline by stacking an S4D layer after the convolutional layer. The convolution layer captures fine-grained local features, while the subsequent S4D module models broader temporal dependencies across the sequence.
    
\begin{figure}
    \centering
    \includegraphics[width=0.8\linewidth]{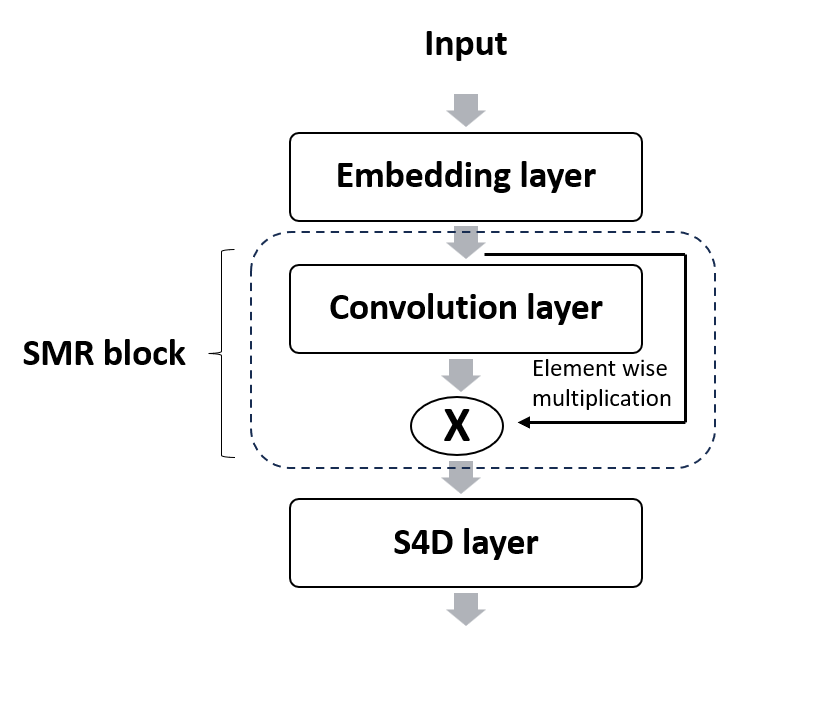}
    \caption{Feature extraction block for SMR + S4D model.}
    \label{fig:smr}
\end{figure}

    \textbf{State Memory Replay(SMR) + S4D model:} The final variant introduces a State Memory Replay (SMR) mechanism \cite{qi2024smrstatememoryreplay} wherein the outputs of the Conv1D layer are multiplied element-wise with the original embedding representations before being passed to the S4D layer, as shown in Figure \ref{fig:smr}. This fusion mechanism improves the ability of the model to recall contextual memory while preserving local structure.

\subsection{Training}
We train all models on the ReVeal dataset \cite{chakraborty2020deeplearningbasedvulnerability}, a real-world benchmark for vulnerability detection. The dataset was constructed from two large-scale open source projects, the Linux Debian kernel and Chromium, covering diverse security issues across operating systems and browsers. It contains 22,734 functions, of which 2,240 are vulnerable and 20,494 are nonvulnerable, formulated as a binary classification task.

To address class imbalance, we used BCEWithLogitsLoss with class weights. All models were trained using Adam Optimizer (learning rate = $1\times10^{-3}$), with dropout 0.5, sequence length capped at 256 tokens, batch size 64, and 10 training epochs. Early stopping was not applied to maintain consistent training duration across runs. The data set was divided into 80\% training, 10\% validation and 10\% testing, and all experiments were implemented in PyTorch and executed on CUDA 11.8.

\subsection{Kernel analysis}
To interpret the internal behavior of the trained models, we performed a kernel-level analysis in both the time domain and the frequency domain.

\subsubsection{Time-domain analysis}
The convolutional and S4D kernels were extracted after training and were visualized in the temporal domain. We evaluated two key properties:
\begin{itemize}[noitemsep,nolistsep]
    \item Peak sensitivity: maximum positive and negative amplitudes, indicating the strongest excitations of the kernel.
    \item Transition sharpness: first-order differences that highlight abrupt changes in kernel response.
\end{itemize}
This analysis reveals whether a kernel is dominated by sharp local transitions or smooth long-range dependencies.

\subsubsection{Frequency-domain analysis}
Frequency-domain analysis was performed using the Fast Fourier Transform (FFT) to obtain the spectral representation of the kernel.
\begin{equation}
\hat{K}(f) = \mathcal{F}{K(t)}.
\end{equation}
Using the fourier transform of the kernel, the normalized Power Spectral Density (PSD) was computed as
\begin{equation}
P(f) = \frac{|\hat{K}(f)|^2}{\sum_{f} |\hat{K}(f)|^2}.
\end{equation}
The following properties were extracted:
\begin{itemize}[noitemsep,nolistsep]
\item \textbf{Dominant frequency:} the frequency with maximum spectral energy.
\item \textbf{Secondary peaks:} modes with magnitudes above 30\% of the dominant frequency.
\item \textbf{Spectral entropy:} measures the concentration or spread of the spectral energy. The spectral entropy is calculated as
\begin{equation}
H = - \sum_{f} P(f)\log P(f).
\end{equation}
\end{itemize}
These measures characterize whether kernels behave as selective narrowband filters or as broadband filters with distributed spectral energy.


\section{RESULTS AND DISCUSSION}
\label{Results}
We evaluated six model architectures on the ReVeal dataset. Table ~\ref{tab:reveal results} summarizes the performance in accuracy, precision, recall, and F1 score. The SMR+S4D model achieved the highest F1 score of 88.03, outperforming the baselines of convolution-only and S4D-only. Separable depth-wise CNNs combined with S4D also performed competitively, demonstrating the benefit of integrating local convolutions with structured state-space representations.

\begin{table}[ht]
\caption{Model performance comparison on the ReVeal dataset for filter size 6. The best result among all architectures is highlighted in bold.}
    \label{tab:reveal results}
    \begin{tabular}{|>{\raggedright\arraybackslash}p{2.5cm} 
                    >{\centering\arraybackslash}p{1.1cm} 
                    >{\centering\arraybackslash}p{1.1cm} 
                    >{\centering\arraybackslash}p{0.9cm} 
                    >{\centering\arraybackslash}p{1cm}|}
    \hline
    Model & Accuracy & Precision & Recall & F1 Score \\
    \hline
     Basic Conv1D       & 87.64 & 87.67 & 87.64 & 87.66 \\
   \hline
    Depthwise Separable Conv  & 87.12 & \textbf{88.08} & 87.12 & 87.57 \\ \hline
    S4D (Standalone)   & 87.07 & 86.97 & 87.07 & 87.02 \\\hline
    Depthwise Separable + S4D & 88.17 & 86.43 & 88.17 & 87.17 \\\hline
    Conv + S4D         & 85.05 & 87.75 & 85.05 & 86.22 \\\hline
    SMR + S4D         & \textbf{88.26} & 87.82 & \textbf{ 88.26} & \textbf{88.03} \\
    \hline
    \end{tabular}
    
\end{table}

\begin{figure*}[!t]
    \centering
    \begin{minipage}{0.48\textwidth}
        \centering
        \includegraphics[width=0.7\linewidth,trim={0cm 1cm 1cm 0cm},clip]{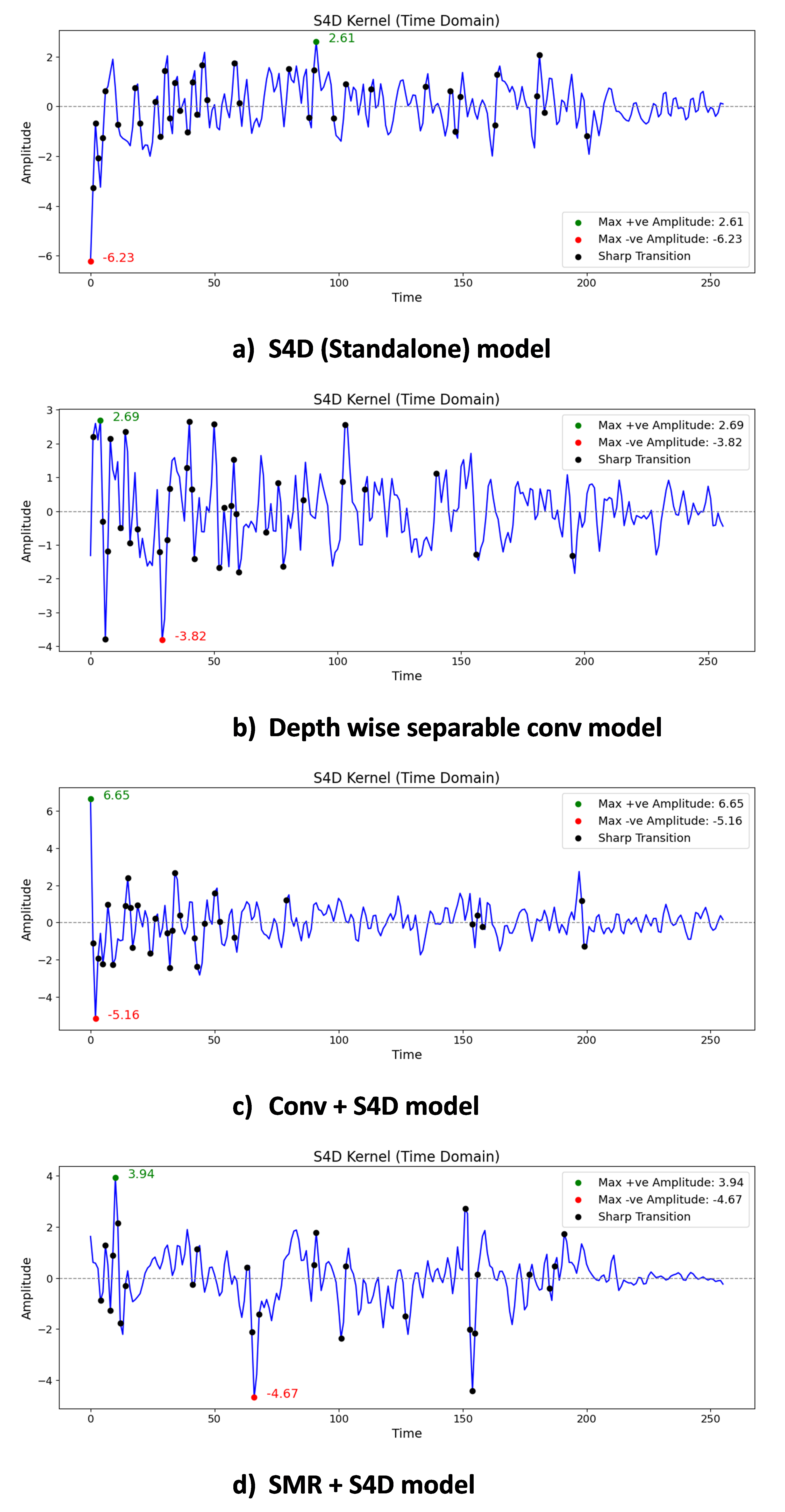}
        \caption{Time-domain impulse responses of S4D kernels across architectures. Green: maximum positive amplitude, red: maximum negative amplitude, black: sharp transitions.}
        \label{fig:time-domain}
    \end{minipage}
    \hfill
    \begin{minipage}{0.48\textwidth}
        \centering
        \includegraphics[width=0.7\linewidth,trim={0cm 4cm 0cm 0cm},clip]{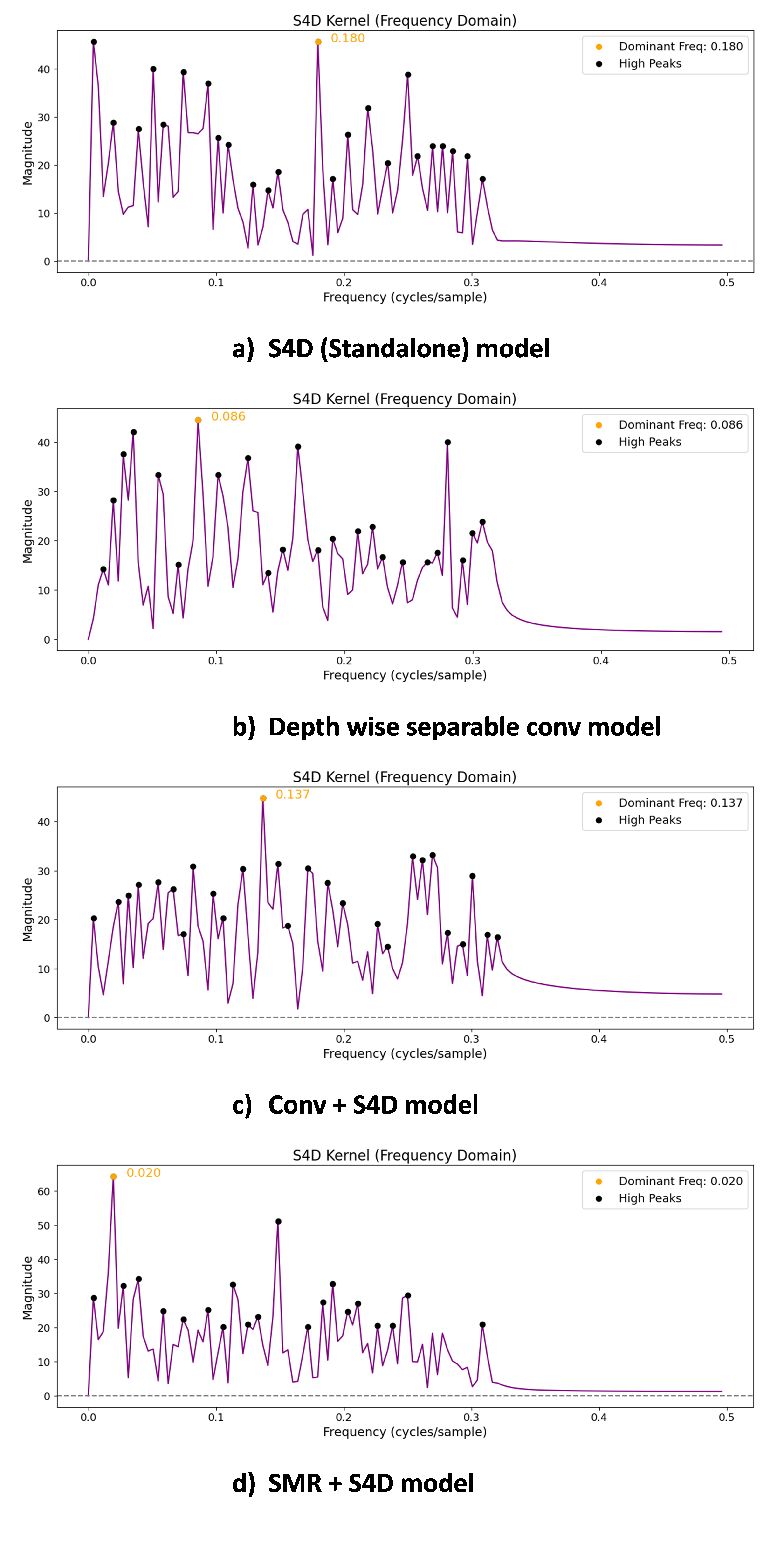}
        \caption{Frequency-domain spectra of S4D kernels across architectures. Orange: dominant frequency; black: secondary peaks ($\geq$30\% of dominant).}
        \label{fig:freq-domain}
    \end{minipage}
\end{figure*}

 We analyzed the learned S4D kernels in both the time and frequency domains to study their dynamic behaviors and complement performance metrics. The time-domain response is shown in Figure \ref{fig:time-domain} and the frequency-domain spectra is shown in Figure \ref{fig:freq-domain}.
The \textbf{standalone S4D} model produced oscillatory kernels with peak amplitudes of approximately $+2.6/-6.2$. Its spectrum was dominated by a mid-frequency component around $0.18$ cycles/sample ($\approx 5$--$6$ token periodicity), with an entropy of $H = 3.94$. This indicates sensitivity to short- to mid-range transitions, consistent with its moderate performance. The \textbf{Depthwise+S4D} variant showed stronger early-sequence modulation ($+2.7/-3.8$) and a lower dominant frequency near $0.086$ cycles/sample ($\approx 12$ tokens). Its entropy of $H = 3.97$, together with multiple mid-frequency peaks, suggests a band-pass profile that captures both local and broader structures. 
The \textbf{Conv+S4D} model exhibited the sharpest temporal responses ($+6.7/-5.2$) and a broadband spectrum with dominant frequency $\approx 0.14$ cycles/sample ($\approx 7$ tokens). Its entropy was the highest ($H = 4.21$), reflecting the distributed spectral energy between frequencies, suggesting that it can engage with both fine-grained syntax and longer-range semantics. 
In contrast, the \textbf{SMR+S4D} kernels had balanced amplitudes ($+3.9/-4.7$) and a very low dominant frequency of $\approx 0.02$ cycles/sample ($\approx 50$ tokens). With the lowest entropy ($H = 3.79$), this model displayed a low-pass profile emphasizing smooth long-range dependencies while suppressing rapid variations, aligning with its superior F1 score of $88.03$.

Thus our analysis reveals that when S4D is preceded by a SMR block, the S4D kernel can capture long-range dependency and the performance of the model improves while the standalone S4D layer cannot capture long-range dependency. This observation is consistent with findings in S4ND \cite{nguyen2022s4ndmodelingimagesvideos}, where band-limiting high-frequency components improved performance. Our analysis suggests that hybrid CNN–SSM architectures can achieve similar band-limiting effects without explicit constraints. Our result is complementary to the previous work \cite{nishikawa2025state}, which proves theoretically that the performance of SSM improves if SSM layer is preceded by feedforward layer.

\section{RELATED WORK}
\label{priorwork}
SSMs have been extended in several directions to improve efficiency, stability, and adaptability. Beyond the original S4 and its diagonalized variant S4D, variants such as S4ND \cite{nguyen2022s4ndmodelingimagesvideos}, StableSSM \cite{wang2024stablessmalleviatingcursememory} and Mamba \cite{gu2024mambalineartimesequencemodeling} have extended the framework to multidimensional data, improved robustness and introduced input-dependent gating. Despite these advances, SSMs face challenges in stability and generalization. Vanilla variants suffer from the curse of memory, where hidden state dynamics collapses into short horizons, preventing effective modeling of long-range dependencies. To overcome this, stability-aware reparameterizations \cite{wang2024stablessmalleviatingcursememory} and initialization scaling methods \cite{liu2024generalizationanalysisoptimizationdesigns} were proposed. Liu et al \cite{liu2024autocorrelationmattersunderstandingrole} emphasize the importance of autocorrelation and eigenvalue placement in initialization, which directly affect memory timescales and learning dynamics. Although these approaches mitigate instability and improve generalization, they do not fully explain the mechanisms that govern kernel behavior.

Previous work in neural networks shows that models often display a spectral bias, learning smooth low-frequency components before fine-grained high-frequency details \cite{rahaman2019spectralbiasneuralnetworks}. Similarly, in convolutional models, the spectrum leakage is observed, where small kernels spread energy broadly across frequencies, while larger kernels concentrate it in low-frequency bands \cite{tomen2021spectralleakagerethinkingkernel, hanin2019finitedepthwidthcorrections}. From a signal processing perspective, long-range-dependent processes concentrate energy in low frequencies and exhibit lower entropy rates \cite{feutrill2021differentialentropyratecharacterisations}. Complementing these findings, S4ND \cite{nguyen2022s4ndmodelingimagesvideos} mitigates spectral leakage by band-limiting noisy high-frequency components, thereby constraining the kernel to low-frequency bands and improving stability and performance in multidimensional modeling tasks. These findings suggest that analyzing SSM kernels through impulse responses and spectral distributions provides valuable insight into whether they capture smooth, global patterns or short, localized fluctuations.
Deep learning models such as CNNs and transformers have been applied to tasks like vulnerability detection \cite{chakraborty2020deeplearningbasedvulnerability}. More recently, CodeSSM \cite{verma2025codessm} demonstrated that SSMs can be promising alternatives for code-related tasks. However, they emphasized on predictive performance, while systematic interpretability analysis of SSM kernels remains limited.

\section{CONCLUSION}
\label{Conclusion}
In this work, we presented the first systematic interpretability study of SSM kernels. By analyzing the S4D model integrated with convolutional layers, we showed that different model architecture result in different filter behavior of the S4D kernel. Our analysis revealed that SMR+S4D exhibit a low-pass bias that emphasizes long-range dependencies resulting in the best performance. The analysis framework presented in the paper can thus be used for understanding how SSM model behaves. The insights can be used to improve architecture of SSM-based models.

\vfill\pagebreak

\bibliographystyle{IEEEbib}
\bibliography{strings,refs}

\begin{thebibliography}{10}

\bibitem{lssl}
Albert Gu, Isys Johnson, Karan Goel, Khaled~Kamal Saab, Tri Dao, Atri Rudra, and Christopher Re,
\newblock ``Combining recurrent, convolutional, and continuous-time models with linear state space layers,''
\newblock in {\em Advances in Neural Information Processing Systems}, A.~Beygelzimer, Y.~Dauphin, P.~Liang, and J.~Wortman Vaughan, Eds., 2021.

\bibitem{verma2025codessm}
Shweta Verma, Abhinav Anand, and Mira Mezini,
\newblock ``Code{SSM}: Towards state space models for code understanding,''
\newblock in {\em The 2025 Conference on Empirical Methods in Natural Language Processing}, 2025.

\bibitem{gu2022parameterizationinitializationdiagonalstate}
Albert Gu, Karan Goel, Ankit Gupta, and Christopher R{\'e},
\newblock ``On the parameterization and initialization of diagonal state space models,''
\newblock in {\em Advances in Neural Information Processing Systems}, Alice~H. Oh, Alekh Agarwal, Danielle Belgrave, and Kyunghyun Cho, Eds., 2022.

\bibitem{gu2022efficientlymodelinglongsequences}
Albert Gu, Karan Goel, and Christopher Re,
\newblock ``Efficiently modeling long sequences with structured state spaces,''
\newblock in {\em International Conference on Learning Representations}, 2022.

\bibitem{qi2024smrstatememoryreplay}
Biqing Qi, Junqi Gao, Kaiyan Zhang, Dong Li, Jianxing Liu, Ligang Wu, and Bowen Zhou,
\newblock ``{SMR}: State memory replay for long sequence modeling,''
\newblock in {\em Findings of the Association for Computational Linguistics: ACL 2024}, Lun-Wei Ku, Andre Martins, and Vivek Srikumar, Eds., Bangkok, Thailand, Aug. 2024, Association for Computational Linguistics.

\bibitem{kim2014convolutionalneuralnetworkssentence}
Yoon Kim,
\newblock ``Convolutional neural networks for sentence classification,'' 2014.

\bibitem{chollet2017xceptiondeeplearningdepthwise}
François Chollet,
\newblock ``Xception: Deep learning with depthwise separable convolutions,'' 2017.

\bibitem{chakraborty2020deeplearningbasedvulnerability}
Saikat Chakraborty, Rahul Krishna, Yangruibo Ding, and Baishakhi Ray,
\newblock ``Deep learning based vulnerability detection: Are we there yet?,'' 2020.

\bibitem{nguyen2022s4ndmodelingimagesvideos}
Eric Nguyen, Karan Goel, Albert Gu, Gordon~W. Downs, Preey Shah, Tri Dao, Stephen~A. Baccus, and Christopher Ré,
\newblock ``S4nd: Modeling images and videos as multidimensional signals using state spaces,'' 2022.

\bibitem{nishikawa2025state}
Naoki Nishikawa and Taiji Suzuki,
\newblock ``State space models are provably comparable to transformers in dynamic token selection,''
\newblock in {\em The Thirteenth International Conference on Learning Representations}, 2025.

\bibitem{wang2024stablessmalleviatingcursememory}
Shida Wang and Qianxiao Li,
\newblock ``Stablessm: Alleviating the curse of memory in state-space models through stable reparameterization,''
\newblock in {\em ICML}, 2024.

\bibitem{gu2024mambalineartimesequencemodeling}
Albert Gu and Tri Dao,
\newblock ``Mamba: Linear-time sequence modeling with selective state spaces,'' 2024.

\bibitem{liu2024generalizationanalysisoptimizationdesigns}
Fusheng Liu and Qianxiao Li,
\newblock ``From generalization analysis to optimization designs for state space models,'' 2024.

\bibitem{liu2024autocorrelationmattersunderstandingrole}
Fusheng Liu and Qianxiao Li,
\newblock ``Autocorrelation matters: Understanding the role of initialization schemes for state space models,'' 2024.

\bibitem{rahaman2019spectralbiasneuralnetworks}
Nasim Rahaman, Aristide Baratin, Devansh Arpit, Felix Draxler, Min Lin, Fred~A. Hamprecht, Yoshua Bengio, and Aaron Courville,
\newblock ``On the spectral bias of neural networks,'' 2019.

\bibitem{tomen2021spectralleakagerethinkingkernel}
Nergis Tomen and Jan van Gemert,
\newblock ``Spectral leakage and rethinking the kernel size in cnns,'' 2021.

\bibitem{hanin2019finitedepthwidthcorrections}
Boris Hanin and Mihai Nica,
\newblock ``Finite depth and width corrections to the neural tangent kernel,'' 2019.

\bibitem{feutrill2021differentialentropyratecharacterisations}
Andrew Feutrill and Matthew Roughan,
\newblock ``Differential entropy rate characterisations of long range dependent processes,'' 2021.

\end{thebibliography}

\end{document}